\documentclass[final,5p,times,twocolumn]{elsarticle}
\usepackage[utf8]{inputenc} 
\usepackage[T1]{fontenc}
\usepackage{amssymb}
\usepackage{graphicx}
\usepackage{picinpar}
\usepackage[fleqn]{amsmath}
\usepackage{url}
\usepackage{colortbl}
\usepackage{soul}
\usepackage{multirow}
\usepackage{pifont}
\usepackage{color}
\usepackage[dvipsnames]{xcolor}
\usepackage{alltt}
\usepackage{hyperref}
\usepackage{enumerate}
\usepackage{siunitx}
\usepackage{breakurl}
\usepackage{epstopdf}
\usepackage{pbox}

\usepackage[ruled, linesnumbered, lined]{algorithm2e}
\usepackage{booktabs}
 
\usepackage{bm}
\usepackage{hyperref}
\usepackage{amssymb}
\usepackage{romannum}
\usepackage{bbm}
\usepackage{booktabs}
\usepackage{array}  
\usepackage{listings}
\usepackage{fancyvrb}
\usepackage{siunitx}
\usepackage[flushleft]{threeparttable}
\usepackage[caption=false,font=footnotesize]{subfig}

\usepackage[normalem]{ulem}

\usepackage[ruled,linesnumbered]{algorithm2e}
\graphicspath{{}}
\DeclareGraphicsExtensions{.eps}
\usepackage{xpatch}
\xpatchcmd{\proof}{\itshape}{\prooflabelfont}{}{}
\newcommand{\prooflabelfont}{\bfseries}

\usepackage[normalem]{ulem}

\makeatletter
\AtBeginDocument{\let\hl\@firstofone}
\AtBeginDocument{\let\hlMath\@firstofone}
\AtBeginDocument{\let\hlTitle\@firstofone}
\makeatother

\journal{Elsevier}

\begin{document}
\begin{frontmatter}
 
\date{}
\title{
Non-Prehensile Tool-Object Manipulation by Integrating LLM-Based Planning and Manoeuvrability-Driven Controls
}

\tnotetext[t1]{
This work is supported in part by the Research Grants Council of Hong Kong under grant C4042-23GF and in part by the National Natural Science Foundation of China (NSFC) under Grant No. 62403211. \\
$\dag$ denotes the corresponding author.
}


\author[1]{Hoi-Yin Lee}
\author[2]{Peng Zhou\textsuperscript{\dag}}
\author[3]{Anqing Duan}
\author[4]{Wanyu Ma}
\author[5]{Chenguang Yang}
\author[1]{David Navarro-Alarcon\textsuperscript{\dag}}

\affiliation[1]{organization={Department of Mechanical Engineering, The Hong Kong Polytechnic University}, addressline={KLN, Hong Kong}
}
\affiliation[2]{organization={School of Advanced Engineering, The Great Bay University}, addressline={Dongguan, China}
}
\affiliation[3]{organization={Department of Robotics, Mohamed Bin Zayed University of Artificial Intelligence}, addressline={Abu Dhabi, UAE}
}
\affiliation[4]{organization={Department of Surgery, The Chinese University of Hong Kong}, addressline={NT, Hong Kong}
}
\affiliation[5]{organization={Department of Computer Science, University of Liverpool}, addressline={Liverpool, UK}
}

\begin{abstract}
The ability to wield tools was once considered exclusive to human intelligence, but it's now known that many other animals, like crows, possess this capability. 
Yet, robotic systems still fall short of matching biological dexterity. 
In this paper, we investigate the use of Large Language Models (LLMs), tool affordances, and object manoeuvrability for non-prehensile tool-based manipulation tasks. 
Our novel method leverages LLMs based on scene information and natural language instructions to enable symbolic task planning for tool-object manipulation. This approach allows the system to convert a human language sentence into a sequence of feasible motion functions.
We have developed a novel manoeuvrability-driven controller using a new tool affordance model derived from visual feedback. This controller helps guide the robot's tool utilization and manipulation actions, even within confined areas, using a stepping incremental approach.
The proposed methodology is evaluated with experiments to prove its effectiveness under various manipulation scenarios.
\end{abstract}


\end{frontmatter}

\section{Introduction}

Being able to use tools is a widely recognised indicator of intelligence across species \cite{stoytchev2007robot, ren2023leveraging}. 
Humans, for instance, have demonstrated mastery of tool use for over two million years.
The ability to use tools is invaluable as it extends an organism's reach and enhances its capacity to interact with objects and the environment \cite{stoytchev2007robot}.
Being able to understand the geometric-mechanical relations between the tools-objects-environments allows certain species (e.g., apes and crows \cite{mccoy2019new}) to reach food in narrow constrained spaces.
The same principles of physical augmentation and its associated non-prehensile manipulation capabilities also apply to robotic systems \cite{jamone2022modelling, liu2022robot}.
For example, by instrumenting them with different types of end-effectors, robots can (in principle) dexterously interact (e.g., push and flip) with objects of various shapes and masses akin to its biological counterpart \cite{huang2023toward, shin2014human, wu2019cooperative} and can be applied to various domains, such as manufacturing \cite{zhang2025dexterous, zhou2024reactive, liu2021sensorless, ma2026large, zhou2025variable}.
However, developing this type of manipulation skill is still an open research problem.
Furthermore, the complexity of planning tool-object manipulation tasks, particularly in coordinating the actions of dual-arm robots, presents significant challenges. To address these complexities, we propose integrating Large Language Models (LLMs) to assist in planning and executing these intricate manipulations, thereby enhancing the robot's ability to perform in diverse scenarios.
\begin{figure}[t]
    \centering
    \includegraphics[width=\linewidth]{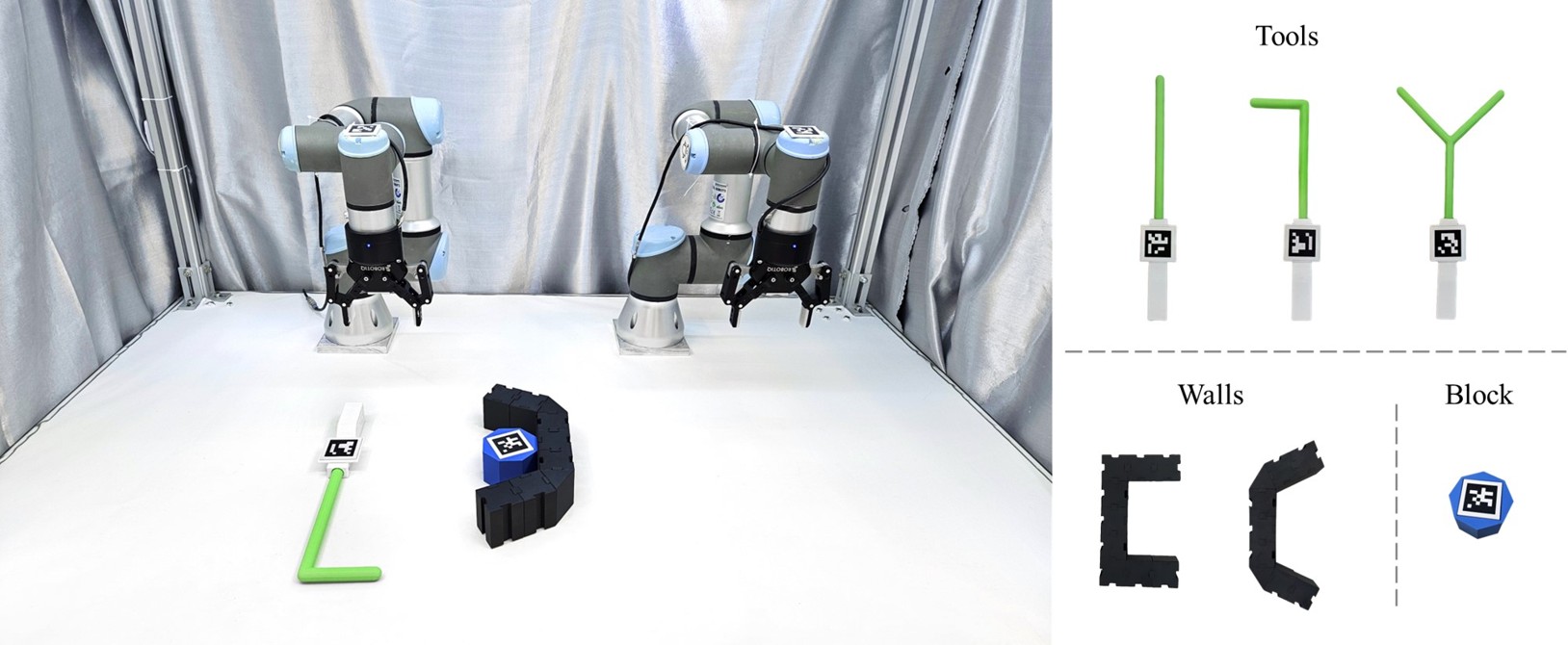}
    \caption{Tool-Object manipulation in a dual-arm robotics system with environmental constraints using the non-prehensile approach.}
    \label{fig:exp_setup}
\end{figure}

Building on the advancements in LLMs, this paper investigates their application alongside tool affordances and object maneuverability for non-prehensile tool-based manipulation tasks. 
Our novel method leverages LLMs based on scene information and natural language instructions to enable symbolic task planning for tool-object manipulation. This approach allows the system to convert a human language sentence into a sequence of feasible motion functions.
We have developed a novel manoeuvrability-driven controller using a new tool affordance model derived from visual feedback. This controller effectively guides the robot's tool utilization and manipulation actions, even in a confined area, using our stepping incremental approach.
The proposed methodology is evaluated with experiments to demonstrate its effectiveness under various manipulation scenarios.

\begin{figure*}[t]
    \centering
    \includegraphics[width=1.0\linewidth]{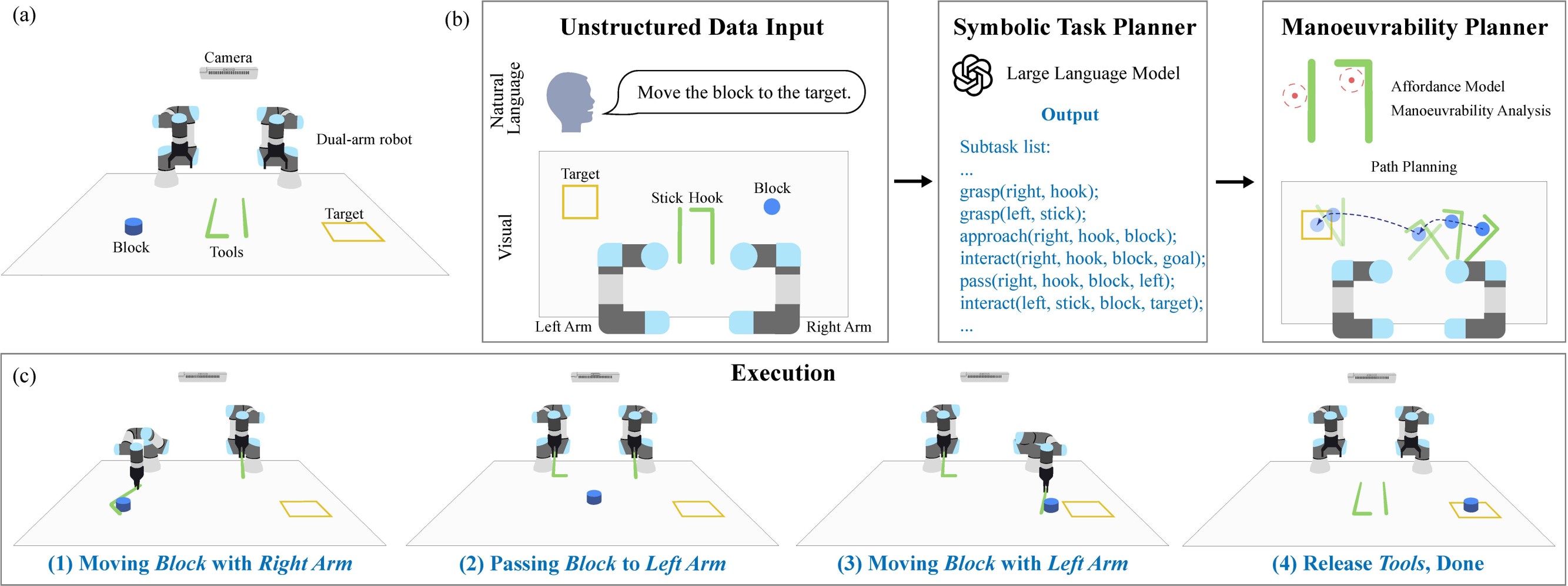}
    \caption{(a) The task environment includes a camera for real-time top-view capturing, a dual-arm robot, tool(s), and a blue manipulandum to be manipulated to the target location. (b) The architecture of our system: Unstructured data input is converted to a subtask list in the symbolic task planner with an LLM, a manoeuvrability-driven planner to compute the tool's manoeuvrability and generate an affordance-oriented motion and path. (c) Execution process of the result given by the system: dual-arm robots take turns pushing the blue manipulandum from one side to another via collaboration. }
    \label{fig:flowchart}
\end{figure*}

\subsection{Related Works}
Effective tool utilisation by a robot involves primarily two aspects: (1) task planning and (2) tool movement \cite{qin2023robot, kroemer2021review, mason2018toward}.
Task planning is typically regarded as a cognitive high-level process in robotics, mainly used for environmental reasoning, task decomposition, allocation of action sequences, etc. \cite{lee2023distributed}.
Task can be decomposed with the integration of learning-based approaches, particularly through the use of reinforcement learning techniques to optimize task planning \cite{kwon2024reinforcement}. Studies have also highlighted the effectiveness of rule-based planning methods, which incorporate predefined heuristics and logical rules to enhance the efficiency of task decomposition in structured environments \cite{veer2023multi}. While rule-based planning is effective for well-defined problems, it can struggle with complex, dynamic environments where the number of rules may become unmanageable. 
However, recent trends have been pushing towards the use of LLMs to leverage the domain knowledge for semantically decomposing and planning the execution of manipulation tasks \cite{wake2023gpt, liu2024delta, rana2023sayplan, qi2026llm, li2025vlm, ahn2022can, huang2022inner, singh2023progprompt, liuvision}.
Some examples of these directions include \cite{ahn2022can, huang2022inner}, which developed an environmental feedback-based system for context-aware improvement planning. 
Leveraging the generative capabilities of LLMs, motion sequences can be generated for robots as demonstrated in \cite{wu2023tidybot, huang2023instruct2act, singh2023progprompt}. 
The combination of traditional motion planners with LLMs has been explored in \cite{wake2023gpt}.
Domain knowledge can be integrated with LLMs to generate a list of motions for navigating a robot in an apartment, as demonstrated in \cite{liu2024delta}. However, the focus primarily remains on independent motions. Motivated by \cite{liu2024delta}, we further consider the dependent motion among arms and tools.

Transitioning from the critical role of task planning, it is evident that effective tool use is inherently tied to understanding the relationship between tools and objects \cite{tsuji2015dynamic}. Indeed, the success of a given tool-object manipulation task largely depends on the appropriate selection of the tool, which necessitates a nuanced comprehension of how tools interact with various objects in their environment.
For example, robots can identify the tool type, potential uses, and contact approaches based on the tool's geometry, see e.g., \cite{qin2023robot, ren2023leveraging}. 
In \cite{wicaksono2016relational}, tool features are learned through observation of the task's effects and experimental validation of feature hypotheses. 
Affordance models are a common technique used for tool feature selection \cite{brawer2020causal, saito2018tool, zech2017computational} and tool classification \cite{tee2018towards, zech2017computational, sinapov2008detecting}.
The relation between tool actions and their effects on objects is explored in \cite{forestier2016modular, sinapov2008detecting}, where robots acquire affordance knowledge through predefined actions (e.g., pull, push, rotate). 
Recently, researchers have also explored the use of LLM in accelerating affordance learning in tool manipulation \cite{ren2023leveraging}. 
Some works have studied tool-based manipulation under constraints and from demonstrations \cite{ding2024task}.
Non-prehensile object manipulation strategies have been used in \cite{imtiaz2023prehensile, selvaggio2023non}.

Building on this foundation of understanding tool-object interactions, it is important to highlight that, despite the advancements in robotic tool use, collaborative tool-based object manipulation by dual-arm systems based on non-prehensile actions remains an underexplored problem.
Notably, the challenge of applying incremental control on the stepping motion of the tool within a confined area has not been well-addressed by previous studies \cite{tsuji2015dynamic, qin2023robot, ren2023leveraging, wicaksono2016relational, ding2024task, zech2017computational, forestier2016modular, ochoa2021impedance, brawer2020causal, saito2018tool, tee2018towards, sinapov2008detecting}. 
Furthermore, most studies have primarily focused on task decomposition for simple object manipulation using LLMs, with tool manipulation being rarely addressed. Dual-arm collaborative manipulation utilizing non-prehensile tools represents a promising area for further exploration. In other words, the integration of LLMs in tool-object manipulation with dual-arm robots remains underexplored. This specific challenge continues to present an open opportunity in the field.

\subsection{Contributions}
To address this research gap, we propose a novel LLM-based manoeuvrability-driven method with the following key contributions:
(1) We develop a geometric-mechanical model that explicitly captures the interaction between tools and objects, enabling accurate representation of their manoeuvrability in various manipulation scenarios;
(2) We introduce a non-prehensile manipulation strategy tailored for tools, allowing efficient object manipulation under various spatial and physical constraints without the need for grasping;
(3) We conduct real-world experiments on a dual-arm robotic system, validating the proposed methodology through performance evaluations and demonstrating its practical applicability in dynamic environments.

Our approach uniquely integrates LLMs to enhance tool-object interactions, enabling robots to interpret and perform complex non-prehensile tasks through natural language instructions. This integration not only improves dynamic adaptability to different manipulation scenarios but also promotes more intuitive human-robot collaboration, increasing the effectiveness of dual-arm tool-object manipulation.

The rest of the manuscript is organised as follows: Sec. \ref{sec:methodology} presents the methodology, Sec. \ref{sec:results} presents the results, Sec. \ref{sec:discussionConclusion} discusses advantages, limitations, and gives final conclusions.

\section{Methodology}\label{sec:methodology}
\subsection{Problem Formulation}
Consider a dual-arm robotic system using a tool to manipulate a block at a far distance (see Fig. \ref{fig:exp_setup}). 
Given the input is a free-form language task $\mathbf L$ (e.g., ``move the block to Point B''), we apply a high-level symbolic planner (i.e., an LLM) to decompose the task into multiple subtasks $\mathbf l_i$, $\mathbf L = \{\mathbf l_1, \mathbf l_2, \dots\}$ where $\mathbf L$ contains a list of pre-defined motion functions $\mathbf l_i$. 

We define a \textit{tool} as a manipulable object that is graspable by a robot, a \textit{manipulandum}  \cite{qin2023robot} as an object (e.g. a block) that is manipulated via a tool, and a \textit{wall} as a static non-manipulable object.
Tool use by robots is challenging as the tools can have various shapes, the environment can be dynamic, and the contact between the tool and the manipulandum may be hard to maintain in a long-horizon task.
In this study, we focus on using the side part of a tool to interact with the \textit{manipulandum}.
Depending on the geometric features of a tool and a wall, the available affordance for manoeuvring a manipulandum may be different. 
Affordance here refers to the available action-effects offered by the tool or the environment.
In this work, we classify affordance into two types: active and passive. Active affordance is given from a manipulable object, i.e. a tool, and it is directly related to the manoeuvrability when driving a manipulandum. A passive affordance is given by a static non-manipulable object. 

To derive our methodology, the following setup assumptions are made: (1) The manipulation motion is planar, (2) the size of the manipulandum is not larger than any one of the segments of the tool, and (3) the manipulandum has a simple, regular geometric shape, such as circular or hexagonal.
Throughout this paper, ``tool-based object manipulation'' is denoted as TOM, and ``tool-based object manipulation under environmental constraints'' is denoted as TOME.
Also, $\mathbf p^{\circ}$ represents the 2D pose of an object $\circ$.
The complete architecture of our method is depicted in Fig. \ref{fig:flowchart}.

\subsection{LLM-Based High-Level Symbolic Task Planner} \label{llm}
To obtain a valid task decomposition for a long-horizon task, the system needs to understand the requirements and generate an executable subtask list. 
We develop a symbolic task planner that takes natural language instructions with scene descriptions as input, and outputs a list of high-level subtasks. 
The list involves the tool selection/sharing between two arms, the sequence to manipulate the tool with the manipulandum, and the interaction between the two arms.  
The model is fine-tuned using approximately 20,000 example data lists, specifically tailored for our non-prehensile tool object manipulation scenario. During the fine-tuning stage, we utilized a program to create 20,000 distinct environmental setups by randomly varying the poses of the robot, tool, block, and target within a finite combination space. 
To ensure data quality and optimality, each generated setup was validated using rule-based filters that enforced logical and task-relevant constraints, ensuring that only feasible and meaningful manipulation scenarios were retained.
This strategy produced a dataset covering both common and rare configurations, enabling the LLM to learn robust mappings between scene layouts and corresponding task plans.
By framing task decomposition as a classification problem, the LLM can effectively associate each setup with a specific list of motion functions.
This design enhances its ability to generate consistent, physically grounded predictions and reduces the likelihood of producing hallucinated or infeasible task sequences.

The system interprets the provided high-level task $\mathbf L$, which can have a structure like ``Please move the blue block to the right-hand side'', ``Can you push the block to the target?'', etc. 
Visual information of the scene is grounded to the system from the observation data $\mathbf o$, where $\mathbf o$ is composed of a series of data points, such as the pose of the block (manipulandum), tools, robots, and walls. 
The system embeds the environmental information with the task instruction to produce a desired configuration requirement, denoted as $\{\mathbf p^{\text{obj}}, \mathbf p^{\text{target}}, \dots \} \leftarrow f(\mathbf L,\mathbf o)$ where $f(\mathbf L,\mathbf o)$ is the embedded result.

The LLM interprets the output of $f(\mathbf L,\mathbf o)$ to generate a subtask list $\{\mathbf l_1, \mathbf l_2, \dots\} \leftarrow f_{\text{llm}}(f(\mathbf L,\mathbf o))$ where $\mathbf l_i$ is a subtask describing the manipulation phase of each robot and corresponds to a high-level robot motion function.
The motion functions are designed to be simple and specify a short-term goal of the concerned object (these functions omit low-level motion commands).  
For simplicity, here we use \textit{m} to represent the manipulandum in the following function definitions. 
We use \texttt{grasp(arm, tool)} for grasping a \textit{tool} with the robot \textit{arm}; 
\texttt{approach(arm, tool, m)} for approaching the location of \textit{m} with \textit{tool} using \textit{arm}; 
\texttt{interact(arm, tool, m, goal)} for moving \textit{m} to the \textit{goal} location with the \textit{tool}; 
\texttt{stepping(arm, tool, m)} for moving \textit{m} out from the bounded area with the \textit{tool} of the \textit{arm} through contact pulsing motions; 
\texttt{pass(arm1, tool, m, arm2)} for passing \textit{m} to another arm's workspace; 
\texttt{release(arm, tool)} for releasing the \textit{tool} back to its original place with the \textit{arm}.

A sample motion task with a dual-arm robot is given as: 
\{\texttt{pass(right, hook, block, left); approach(left, stick, block); interact(left, stick, block, target)}; \dots \} $\leftarrow f_{\text{llm}}(f(\mathbf L,\mathbf o))$
where both arms take turns manipulating the block.
The right arm passes the block to the left by pushing it to an area where both arms can reach it.
The left arm approaches the block with a stick and manipulates the block to the target.  
To this end, the symbolic task planner converts the unstructured data to a series of motion functions, including robot motion, tool planning, manipulation sequence, and collaboration.

\subsection{Visual Affordance Model} \label{affordanceModel}

\begin{figure}[t]
    \centering
    \includegraphics[width=1\linewidth]{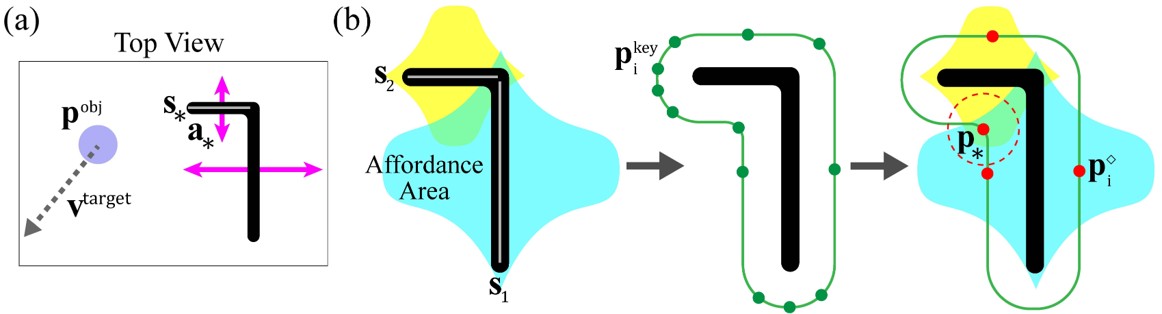}
    \caption{(a) Affordance vectors are shown in pink arrows. Grey arrow is $\mathbf{v}^{\text{target}}$ and the desired affordance vector is denoted as $\mathbf a_\ast$. (b) shows the manoeuvrability analysis flowchart: affordance area is visualised with the Gaussian function in yellow and blue; expand and downsample the tool's shape to get key points $\mathbf P^{\text{key}}$ (green colour dots); combine the affordance area with the key points $\mathbf P^{\text{key}}$ to get the non-redundant points $\mathbf P^{\diamond}$ (red dots), and combine the affordance $\mathbf a_\ast$ found in (a) to obtain the position for the manipulandum to be at with the tool (labelled as $\mathbf p_\ast$ with a red dot) and the highest manoeuvrability region is shown with a dashed red circle.}
    \label{fig:tool_flowChart}
\end{figure}

Tools can have various shapes and complex structures. 
In this paper, we focus on the following tool geometries: a stick, an L-shaped hook, and a Y-shaped hook.
Affordances are related to the geometric features of a tool.
To analyse the possible affordances, we divide the tool into smaller segments (i.e. a line), and denote them as $\mathbf S = \{\mathbf s_1, \mathbf s_2, \dots, \mathbf s_n\}$ where $\mathbf s_i$ and $\mathbf s_{i+1}$ are segments next to each other.
We compute the normal vectors of the segment at the middle point and scale them by half of the segment's length. This is done to weigh the affordance effect this region carries.
There are two affordance vectors per segment $\mathbf s_i$, each pointing in opposite directions, as depicted in Fig. \ref{fig:tool_flowChart}(a). 
Let us define $\mathbf A = \{\mathbf a_1, \mathbf a_2, \dots, \mathbf a_{2n}\}$ as the structure that contains all the affordance vectors $\mathbf a_i$, for $n$ as the number of segments.

To determine which affordance vector $\mathbf a_i$ will be used to interact with the manipulandum, we compare the similarity between $\mathbf a_i$ and the vector from the manipulandum's position to the target point $\mathbf v^{\text{target}}$ by:
\begin{equation} \label{vecSimilarity}
    \theta_i = \cos^{-1}{\left(\frac{\mathbf v^{\text{target}} \cdot \mathbf a_i}{\|\mathbf v^{\text{target}}\| \|\mathbf a_i\|}\right)}
\end{equation}
where $\theta_i$ is the similarity score.
The optimal affordance vector $\mathbf a_{\ast}$ and its according segment $\mathbf s_\ast$ are found by:
\begin{equation}
    \mathbf a_{\ast} = \mathop{\arg\min}_{\mathbf a} (\Theta) \ \ \text{for} \ \Theta = \{\theta_1, \theta_2, \dots \}
\end{equation}
where the vector with the minimum similarity score is the optimal affordance vector.

\subsection{Manoeuvrability Analysis} \label{manAnalysis}
A tool can push the manipulandum from the side, from the tip, or from other areas. 
However, the relative location of the manipulandum with respect to the tool affects its manoeuvrability. 
In other words, the affordance provided by the tool is proportional to manoeuvrability. 
Consider using a rotating stick to push an object with its end tip.
In this situation, the tool may lose contact with the manipulandum as it rolls outwards; hence, the manoeuvrability of this point is low. 
On the other hand, the midpoint of the stick has a high manoeuvrability, which proportionally decreases as the contact point is further away from the midpoint.
This behaviour can be modelled with a Gaussian function, where its centre is the segment's centre and the peak height is half the segment's length, see Fig. \ref{fig:tool_flowChart}(b). 
We refer to this region as an affordance area.

All the pixels in the affordance area of $\mathbf s_i$ are set to $1$ in an image frame $\mathbf I_i$ and the rest to $0$, which creates a binary image; This process is repeated for all segments.
All binary images are then summed as:
\begin{equation}
    \hat{\mathbf I} = \sum_{i=1}^{n} \mathbf I_i, \ \ \
    [\mathbf I]_{x,y}  = 
    \begin{cases}
        1,& \text{if it is an affordance area}\\
        0,& \text{else}
    \end{cases}
\end{equation}
where $n$ is the number of segments. 
The affordance of the tool segment is quantified with the (normalised) manoeuvrability matrix: $\mathbf M =  \hat{\mathbf I} / \hat I_{\text{max}}$, for $\hat I_{\text{max}}$ as the maximum value in $\hat{\mathbf I}$. 

Tool regions with high values in the image $\mathbf M$ reflect a high manoeuvrability.
These computed manoeuvrability values are useful to determine the location where the tool interacts with the manipulandum. 
To determine the centre of the object, we then expand the contour of the tool by the object's radius $r^{\text{obj}}$. 
This contour is downsampled with the Ramer-Douglas-Peucker algorithm, then parameterised with the spline fitting technique.
To extract key features of the tool geometry, we use a sliding window strategy to examine a small number of neighbouring points. 
Let $\mathcal{C}$ be the contour of the tool expanded by $r^{\text{obj}}$. The key features of the tool geometry are extracted using the following equation:
\begin{equation}
\mathcal{F} = \{ p \in \mathcal{C} | \kappa(p) > \kappa_{\text{thresh}} \}
\end{equation}
where $\mathcal{F}$ is the set of feature points, $p$ represents a point on the parameterized contour $\mathcal{C}$, $\kappa(p)$ is the curvature of the point $p$, and $\kappa_{\text{thresh}}$ is a predefined curvature threshold.
If there exists a point where its curvature is larger than a threshold in the local neighbourhood, we consider this point as one of the feature points. 

To compute the minimal number of key points (denoted as $\mathbf P^{\text{key}} = \{\mathbf p^{\text{key}}_1, \mathbf p^{\text{key}}_2, \dots\}$) that capture the highest manoeuvrability among feature points, we use the density‐based clustering algorithm. 
By integrating the affordance areas we obtained earlier, we can filter out some redundant key points.
For example, if there exists a point $\mathbf p^{\text{key}}_i$ located outside the affordance area (visualised in Fig. \ref{fig:tool_flowChart}(b)), we consider this point as redundant.
All the non-redundant points are then grouped into $\mathbf P^\diamond = \{\mathbf p^\diamond_1, \mathbf p^\diamond_2, \dots\}$.
To find the point in $\mathbf P^\diamond$ with the highest manoeuvrability (defined as $\mathbf p_\ast$), we use the manoeuvrability matrix $\mathbf M$ and distance between $\mathbf p^\diamond_i$ and $\mathbf a_\ast$ as described in the metric below:
\begin{equation}
   \mathbf p_\ast = \mathop{\arg\min}\limits_{\mathbf p^\diamond_i}((1-[\mathbf M]_{\mathbf p^\diamond_i}) + \|\mathbf p^\diamond_i - \mathbf a_\ast\|)
\end{equation}
where $[\mathbf M]_{\mathbf p^\diamond_i}$ denotes to the image value of $\mathbf M$ at point $\mathbf p^\diamond_i$. 
The region with the highest manoeuvrability is defined as the circle (with object radius) centred at  $\mathbf p_\ast$. (see Fig. \ref{fig:tool_flowChart}(b))

\begin{figure}[t]
    \centering
    \includegraphics[width=1\linewidth]{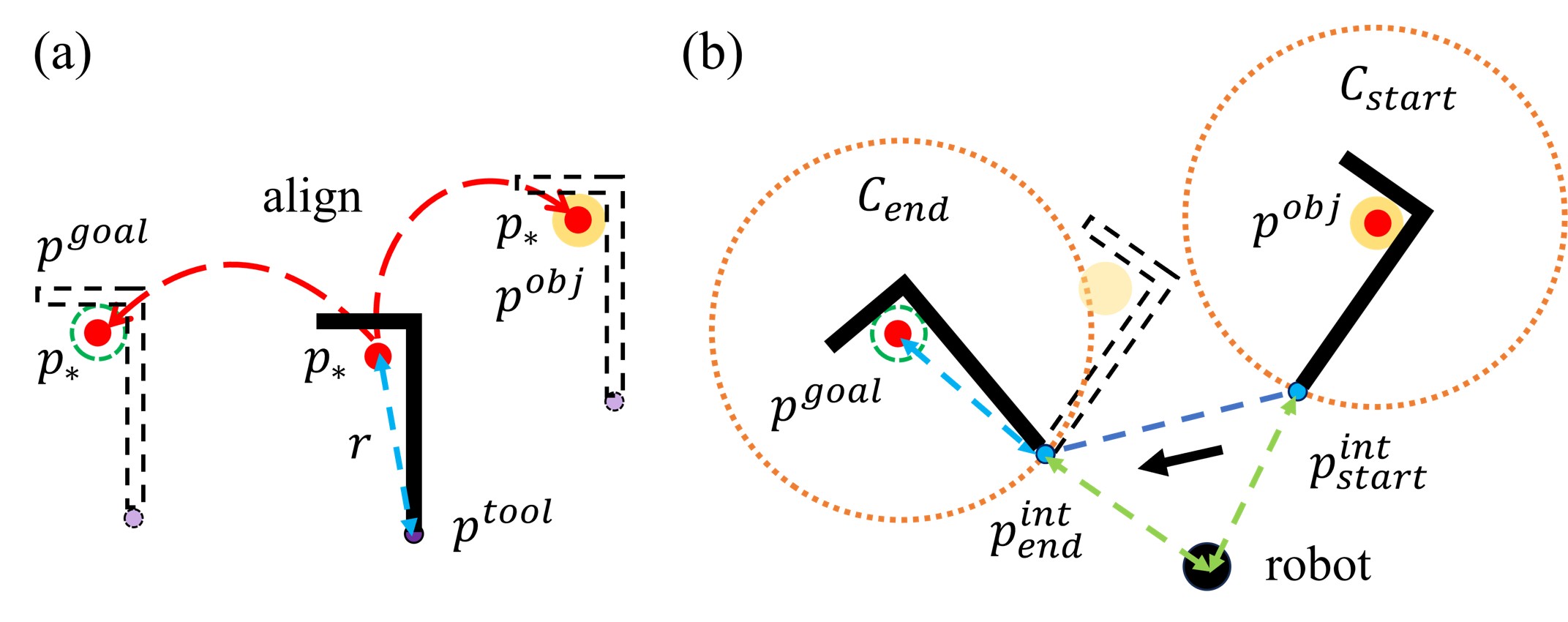}
    \caption{(a) The tool is virtually aligned to the current object and the goal location, with $\mathbf p_\ast = \mathbf p^{obj}$ and $\mathbf p_\ast = \mathbf p^{goal}$. (b) The light blue dashed line is the radius of the orange circle $\mathbf C_{\text{start}}$ and $\mathbf C_{\text{end}}$, which equals the distance between $\mathbf p^{tool}$ and $\mathbf p_\ast$. The tool moves from $\mathbf p^{\text{int}}_{\text{start}}$ to $\mathbf p^{\text{int}}_{\text{end}}$ by following the dark blue dashed trajectory line.}
    \label{fig:path}
\end{figure}

\subsection{Manoeuvrability-Oriented Controller}
The subtask ``$\mathtt{interact}$" triggers the robot to use the selected tool to drive the manipulandum towards the desired location. 
In this section, we derive our method to perform this type of motion assuming that the tool approaches the object and is going to make contact with it in the subtask ``$\mathtt{interact}$".

\subsubsection{Initial and Final Poses}
The tool's pose corresponds to its grasping configuration, which coincides with the robot end-effector's pose when the robot grasps the tool (see Fig. \ref{fig:path}).
$\mathbf{p}^{\text{tool}}$ denotes the tool's grasping point ($x,y$ coordinates) when it has not come in contact with the object.
To construct a trajectory for tool-based object transport, we need to find out the tool's desired initial and final poses for the subtask ``$\mathtt{interact}$".
We first define these poses (which include the orientation) of the chosen tool as $\mathbf{p}^{\text{int}}_{\text{start}}$ and $\mathbf{p}^{\text{int}}_{\text{end}}$ respectively. 

To efficiently move the object, we propose a method that reduces the travel distance while ensuring continuous contact. 
In the first contact, we align the highest manoeuvrability point $\mathbf p_\ast$ of the tool to the object's centre $\mathbf p^{\text{obj}}$, where $\mathbf p_\ast = \mathbf p^{\text{obj}}$.

The motion trajectory of a tool, moving along the z-axis of the object's centre without displacing it can be described as a circular trajectory with the centre $\mathbf p^{\text{obj}}$ and radius $r$, where $r = \|\mathbf p_\ast - \mathbf{p}^{\text{tool}}\|$.
The trajectories for the initial and final configurations are represented as $\mathbf C_{\text{start}}$ and $\mathbf C_{\text{end}}$ (see Fig. \ref{fig:path}(a)).

The possible location for $\mathbf p^{\text{int,x,y}}_{\text{start}}$ will be lying on $\mathbf C_{\text{start}}$ and can be determined by finding a point on $\mathbf C_{\text{start}}$ which is the closest point to the robot (the distance is indicated with a light green dashed line in Fig.\ref{fig:path}(b)).
Based on the tool's geometry, we can determine the orientation of the initial pose $\mathbf p^{\text{int}}_{\text{start}}$; The same approach applies to $\mathbf p^{\text{int}}_{\text{end}}$.

\subsubsection{Motion Strategy}
To stably move from $\mathbf p^{\text{int}}_{\text{start}}$ to $\mathbf p^{\text{int}}_{\text{end}}$, the following motion strategy is implemented to achieve the task: First, the robot aligns $\mathbf p_\ast$ with $\mathbf p^{\text{obj}}$ and matches $\mathbf p^{\text{tool}}$ with $\mathbf p^{\text{int}}_{\text{start}}$ with the following equation:
\begin{equation}
    \mathbf p^{\text{tool}} = \text{argmin}_{\mathbf{p}}(f(\mathbf p)) + ||\mathbf p - \mathbf p^{\text{int}}_{\text{start}}||)
\end{equation}
where the coordinates of $\mathbf{p}^{\text{tool}}$ can be determined by finding a point $\mathbf p = (x,y)$ where it minimizes the distance between $(\mathbf p_\ast, \mathbf p^{\text{obj}})$ with $f(\mathbf p)$ and $(\mathbf{p}^{\text{tool}}, \mathbf p^{\text{int}}_{\text{start}})$; 
then translates along the $x$ and $y$ axes until it reaches $\mathbf p^{\text{int,x,y}}_{\text{end}}$ with $k_\text{int}(\mathbf p^{\text{int,x,y}}_{\text{end}} - \mathbf p^{\text{tool}})$, where $k_\text{int}$ is determined empirically; 
lastly, the tool is rotated to align with the orientation of $\mathbf p^{\text{int}}_{\text{end}}$.

\begin{figure}[t]
    \centering
    \includegraphics[width=1\linewidth]{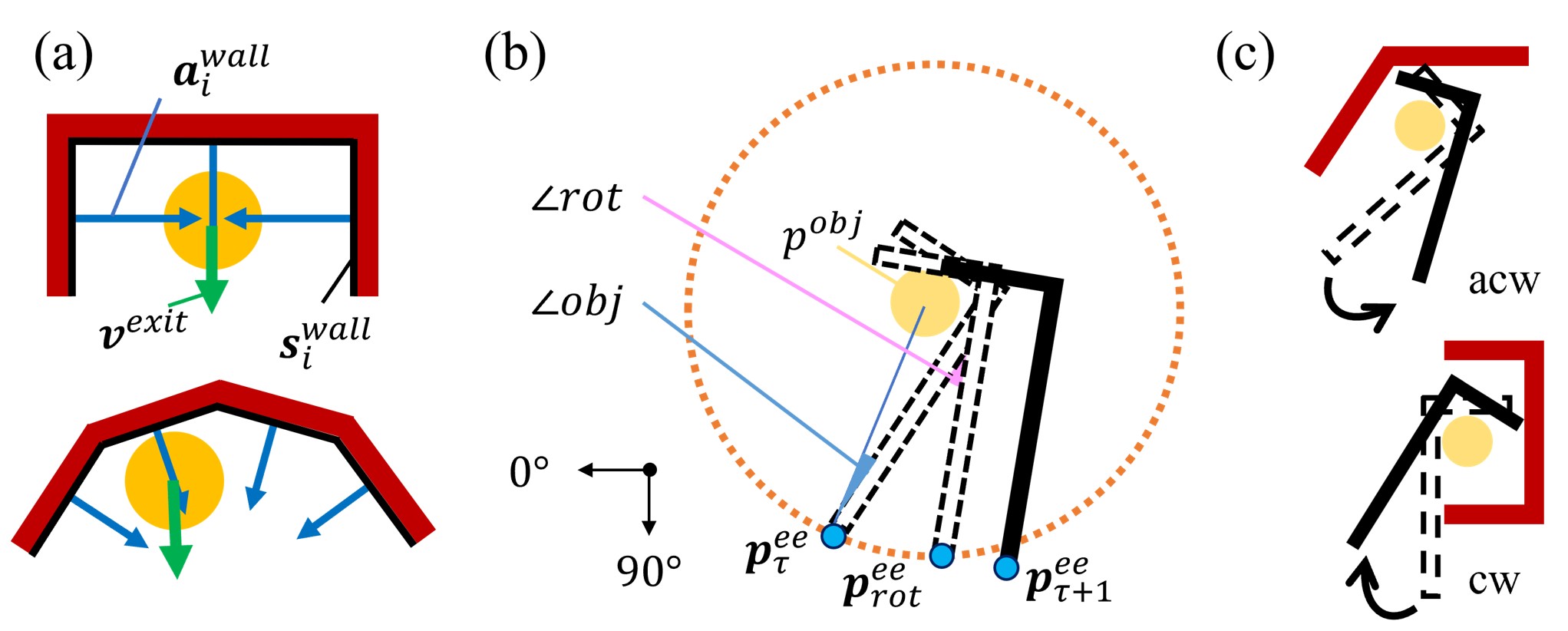}
    \caption{(a) Walls are in red with the segment of the wall $\mathbf s^{\text{wall}}_i$ highlighted in black; blue arrows are the passive affordance vector and green arrows indicate the moving direction of $\mathbf v^{\text{exit}}$. 
    (b) The tool pose moves from $\tau$ to $\tau+1$ by rotating with $\angle{\text{rot}}$ and translating linearly to $\mathbf p^{\text{ee}}_{\tau+1}$. (c) Rotation direction of a tool: anti-clockwise and clockwise direction.}
    \label{fig:rotation_direction}
\end{figure}

\subsection{Application with Environmental Constraints} \label{envConstraints}
When moving an object across a table, we may encounter constraints from the environment, such as walls. These constraints restrict the potential movement directions of the object. 
Formally, a constrained area can be defined by a series of points where more than one axis of freedom of the manipulandum motion may be restricted. In this section, we focus on the motion triggered by the subtask `$\mathtt {stepping}$'.

Consider the manipulandum is tightly confined within a concave-shaped wall, as shown in Fig. \ref{fig:rotation_direction}(a), with an unknown exit and assume that the tool can enter the constrained area. 
To move the manipulandum out of the bounded area with a small movement space, we determine the direction from the manipulandum to the exit by considering the overall affordance of the wall boundary. We denote this direction vector as $\mathbf{v}^{\text{exit}}$, and its magnitude is defined as the minimum travel distance for the manipulandum.
Consider the inner edge of the wall as a segment $\mathbf s^{\text{wall}}_i$ where $i =\{1, \dots, n^{\text{wall}}\}$ and $n^{\text{wall}}$ is the number of the wall segment.
The affordance of a wall is passively provided and is defined as $\mathbf a^{\text{wall}}_i$ with the model shown in Sec. \ref{affordanceModel}. 
The passive affordance vector is the normal vector of $\mathbf s^{\text{wall}}_i$ located in the middle with the direction pointing towards the constrained area. Its magnitude is scaled to half of $\mathbf s^{\text{wall}}_i$ as the manipulandum is generally not receiving any affordance from a wall segment based on our visual affordance model.
The moving direction for the manipulandum to the exit can be obtained by the following equation: 
\begin{equation}
    \mathbf v^{\text{exit}} = \sum_{i=1}^{n^{\text{wall}}} \mathbf a^{\text{wall}}_{i} + \mathbf p^{\text{obj}}
\end{equation}
where $\mathbf v^{\text{exit}}$ integrates all passive wall affordance vectors $\mathbf a^{\text{wall}}_i$ with the current position of the manipulandum, see \ref{fig:rotation_direction}(a).

Given that only part of the tool can enter the confined area, our primary focus is the tip of the tool. 
The segment connecting of the tool's tip is denoted as $\mathbf{s}^{\text{tip}}$, with its corresponding affordance vector denoted as $\mathbf{a}^{\text{tip}}$. The desired rotation angle of the end pose of $\mathbf{a}^{\text{tip}}$ is the angle of $\mathbf{v}^{\text{exit}}$.

The highest manoeuvrability region can be obtained by treating $\mathbf v^{\text{exit}}$ as the target vector $\mathbf v^{\text{target}}$, $\mathbf a^{\text{tip}}$ as the desired affordance $\mathbf a_\ast$, and assuming the tool is rotated such that $\mathbf a^{\text{tip}} = b  \mathbf v^{\text{exit}}$ with $b>0$ as a scaling factor.
We first align $\mathbf s^{\text{tip}}$ to the first segment of the wall (i.e. $\mathbf s_1$), with $\mathbf p^{\text{obj}}$ inside the highest manoeuvrability region of the tool.
The tool approaches the object and maintains contact with the manipulandum by minimising the distance $\|{\mathbf p}_\ast - \mathbf p^{\text{obj}}\|$.

To move in the limited area while interacting with the manipulandum, we employ a stepping approach to manipulate the manipulandum in the confined area.
As the possible movement area is small and highly restricted, an incremental pulsing motion is adopted to make small adjustments with high accuracy motion control to the tool and the manipulandum.
Inspired by the animal manipulation study in \cite{mccoy2019new} (where a crow uses a tool to get the food from the box slot by rotating and dragging the tool outwards), we adopt a similar approach to retrieve the object from confined spaces. 
This strategy continuously alternates between ``repositioning'' the tool and incremental ``rotation-dragging'' the object towards the exit until it can be fully extracted as depicted in Fig. \ref{fig:rotation_direction}.

We define ``repositioning'' as moving the tool closer to the object and realigning $\mathbf p_\ast$ with $\mathbf p^{\text{obj}}$ by $k$ amount. The value of $k$ is determined empirically, representing the spatial offset between the tool and the object. A larger $k$ allows a wider clearance before contact, while a smaller $k$ brings the tool closer, increasing precision but also the risk of collision.
In ``rotation-dragging'', the tool maintains contact with the manipulandum when it rotates by a certain angle as $\angle{\text{rot}}$ shown in Fig. \ref{fig:rotation_direction}(b) and moves outwards by extending $\overrightarrow{\mathbf p^{\text{ee}}_\tau \mathbf p^{\text{ee}}_{\text{rot}}}$ by a $w>0$ amount. 
If $\angle{\text{rot}}$ or $w$ are excessively large, the tool may jam or cause damage in the constrained area; conversely, values that are too small reduce efficiency by requiring more iterations to complete the manipulation. 
It is a trade-off between maintaining stability and achieving motion efficiency.

$\tau$ is an action step variable and is incremented by $1$ if an action (reposition/rotation-dragging) is fulfilled (i.e. $\tau = 0,1,2,\dots$). 
To control the change of action, a step function (denoted as $u(\tau)$) is implemented as a trigger with the step variable $\tau$.
This kind of non-prehensile crow-inspired behaviour can be unified and modelled as:
\begin{equation} 
\begin{aligned}
    \mathbf p^{\text{ee}}_{\tau+1} &= 
                        \begin{bmatrix}
                            \mathbf p^{\text{ee},x}_\tau\\
                            \mathbf p^{\text{ee},y}_\tau\\
                            \phi_{\tau}
                        \end{bmatrix} +
                         u(\tau)
                        \begin{bmatrix}
                            k(\mathbf p^{obj,x}_\tau - \mathbf p^x_\ast)\\
                            k(\mathbf p^{obj,y}_\tau - \mathbf p^y_\ast)\\
                            0
                        \end{bmatrix}\notag \\ 
                        &\quad +
                        u(\tau + 1)
                        \begin{bmatrix}
                            w(\mathbf p^{\text{obj},x}_\tau - r\cos(\phi_\tau) - \mathbf p^{\text{ee},x}_\tau)\\
                            w(\mathbf p^{\text{obj},y}_\tau + r\sin(\phi_\tau) - \mathbf p^{\text{ee},y}_\tau)\\
                            f(\phi_{\tau+1})
                        \end{bmatrix}
\end{aligned}
\label{eq:nonPrehensile}
\end{equation}

\begin{equation}
    u(\tau) = 
    \begin{cases}
        0,& \text{if } \tau \text{ is odd}\\
        1,& \text{if } \tau \text{ is even}
    \end{cases}
\end{equation}

where $\mathbf p^{\text{ee}}_{\tau+1}$ is the next target pose of the end-effector at the action step $\tau+1$ for the affordance vector $\mathbf a^{\text{tip}}$ not parallel to $\mathbf v^{\text{exit}}$, such that $\mathbf a^{\text{tip}} \neq b  \mathbf v^{\text{exit}}$. 
The angle of the tool at $\tau+1$ (denoted as $\phi_{\tau+1}$) depends on the rotational direction (see Fig. \ref{fig:rotation_direction}), that $\phi_{\tau+1}$ is computed as
\begin{equation}
    f(\phi_{\tau+1}) = \left\{\begin{aligned}
    & - \angle{\text{obj}} - \angle{\text{rot}},~~ \text{if direction is anti-clockwise} \\
    & - \phi_{\tau} +\pi - \angle{\text{obj}} - \angle{\text{rot}},~~ \text{otherwise}
    \end{aligned}\right.
\end{equation}
where $\phi_{\tau}$ is the tool's angle at the action step $\tau$, $\angle{\text{obj}}$ is the angle between the manipulandum, grasping point, and a tool's keypoint, $\angle{\text{rot}}$ is the amount of angle to rotate.

\section{Results}\label{sec:results}

To evaluate the proposed framework in terms of accuracy, robustness, and practical feasibility, approximately 200 experiments are conducted using a dual-arm UR-3 robotic system.
The fine-tuning of the large language model (GPT-4o-mini) is performed in the cloud on GPU-enabled servers, and during deployment, the robotic system accesses the trained model through a secure API connection for inference.
Three types of tools are selected, which are a stick, an L-shaped hook, and a Y-shaped hook (see Fig.\ref{fig:exp_setup}).
Different combinations of these tools were evaluated under diverse movement directions and task objectives.
Various masses of the manipulandum are tested, but due to the minimal impact on the vision-based controller, mass is excluded from this section.
The experimental tasks covered a wide range of scenarios, including close-range manipulation with single and multiple tools, long-horizon (single and tool-sharing) operations, and manipulation within constrained environments.
An Intel RealSense D415 captures the images of the whole process. Data is passed to a Linux-based computer with the Robot Operating System (ROS) for image processing and robot control.
Aruco markers are used for providing accurate pose tracing in real time. 
The average inference time is approximately 0.158 seconds for tool analysis and around 1.51 seconds for LLM processing. Since these operations are completed prior to robot movement, their latency had minimal influence on overall system responsiveness.

These experiments include validating the task decomposition performance in a single and dual-arm robot setup, the robustness of the affordance and manoeuvrability model in various shapes of tools, and evaluating the overall performance.

\begin{figure}[t]
\centering
  \includegraphics[width=1\linewidth]{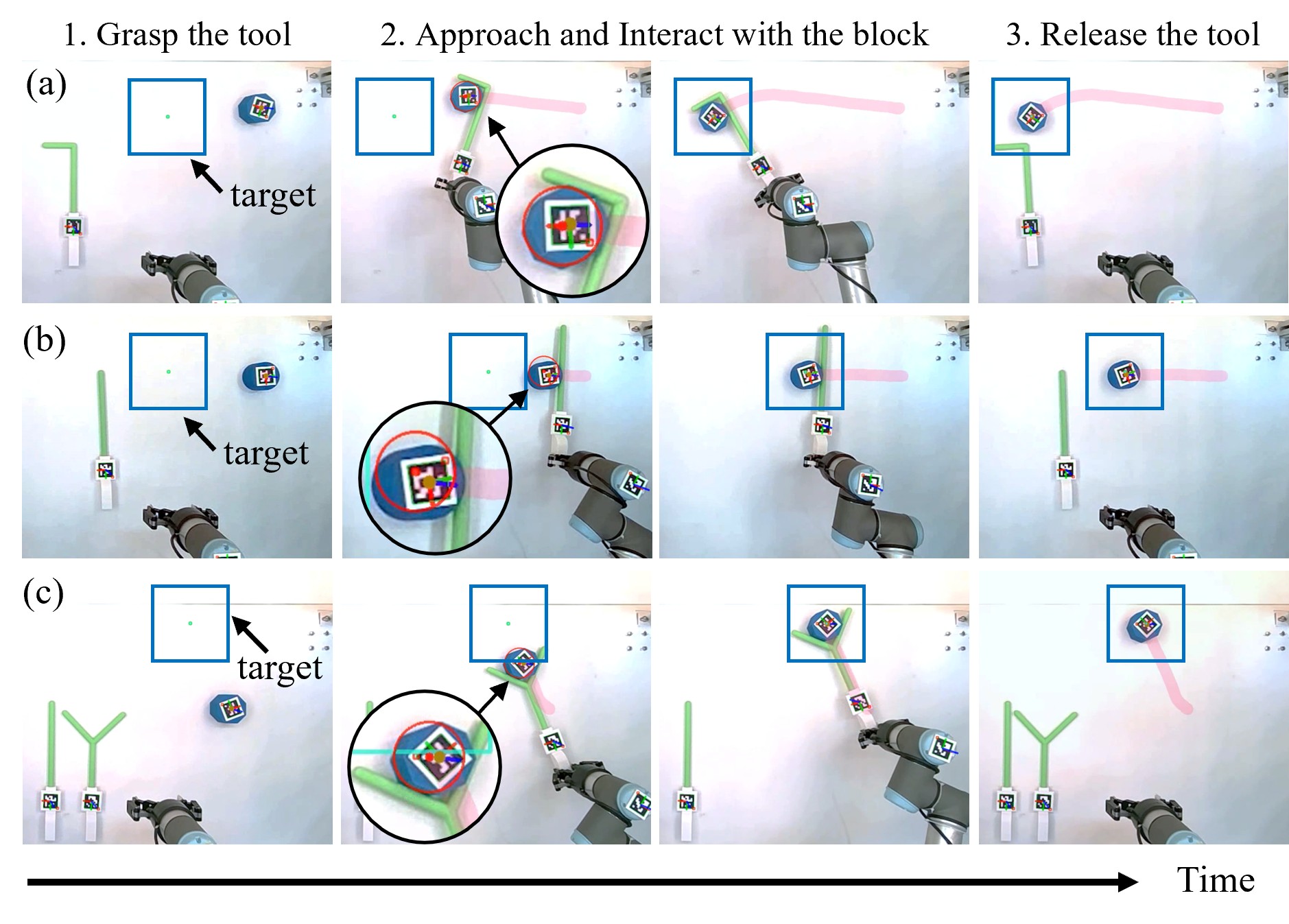}
  \caption{Single-arm robot with a single tool: moving the manipulandum (a) right to left with a hook, (b) right to left with a stick, and (c) bottom to top with a Y-shaped tool. The red line shows the manipulandum's trajectory, while the red circle indicates the highest maneuverability point.}
  \label{fig:exp1}
\end{figure}

\begin{figure}[t]
    \centering
    \includegraphics[width=1\linewidth]{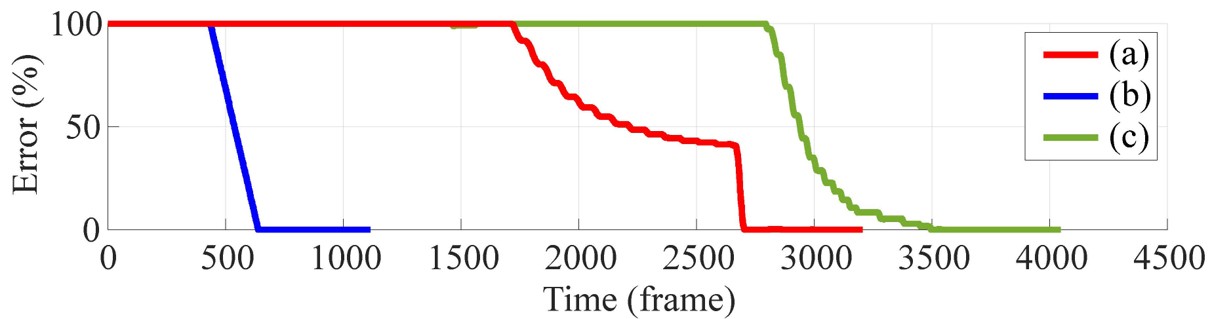}
    \caption{Evolution of the minimisation process of the error between the current object position and the target for the tasks shown in Fig. \ref{fig:exp1}.}
    \label{fig:exp1Error}
\end{figure}

\subsection{Single-Arm Robot}
We first evaluate the task decomposition performance of LLM. For that, a tool and a blue manipulandum are placed on the table with the target given as shown in Fig. \ref{fig:exp1}. The task is to manipulate the manipulandum within a close distance, which is sufficient for a single-arm robot. 
The embedded information, which contains the task, the environment, and the geometry of the tool, is passed to the LLM. In the experiment shown in Fig. \ref{fig:exp1}(a), the robot executes the subtasks generated by the high-level symbolic task planner, which include: 
\texttt{grasp(right, hook); 
approach(right, hook, block); 
interact(right, hook, block, target); 
release(right, hook)}.
The right arm first moves and grasps the hook, then moves the block to the target, and lastly releases the tool back to its original place. 

In a non-single tool scenario, where two tools are available on the desk as shown in Fig. \ref{fig:exp1}(c), the task planner selects the nearest tool based on the embedded information to push the block towards the target.
The experiment showcases the application of the proposed affordance and manoeuvrability model in locating the highest manoeuvrability region for manipulandum transportation.
During the manipulation stage, the manipulandum is kept within the highest manoeuvrability region (indicated with a red circle in Fig. \ref{fig:exp1}) to receive affordance effectively from the tool. 
The minimisation of the error between the $\mathbf p^{\text{obj}}$ and the $\mathbf p^{\text{target}}$ for each experiment is shown in Fig. \ref{fig:exp1Error}. These results corroborate that the proposed method can be used to actively drive a robot to manipulate an object via a tool.

\subsection{Dual-Arm Robot with Long-Horizon Task}
\begin{figure*}[t]
    \centering
    \includegraphics[width=1.0\linewidth]{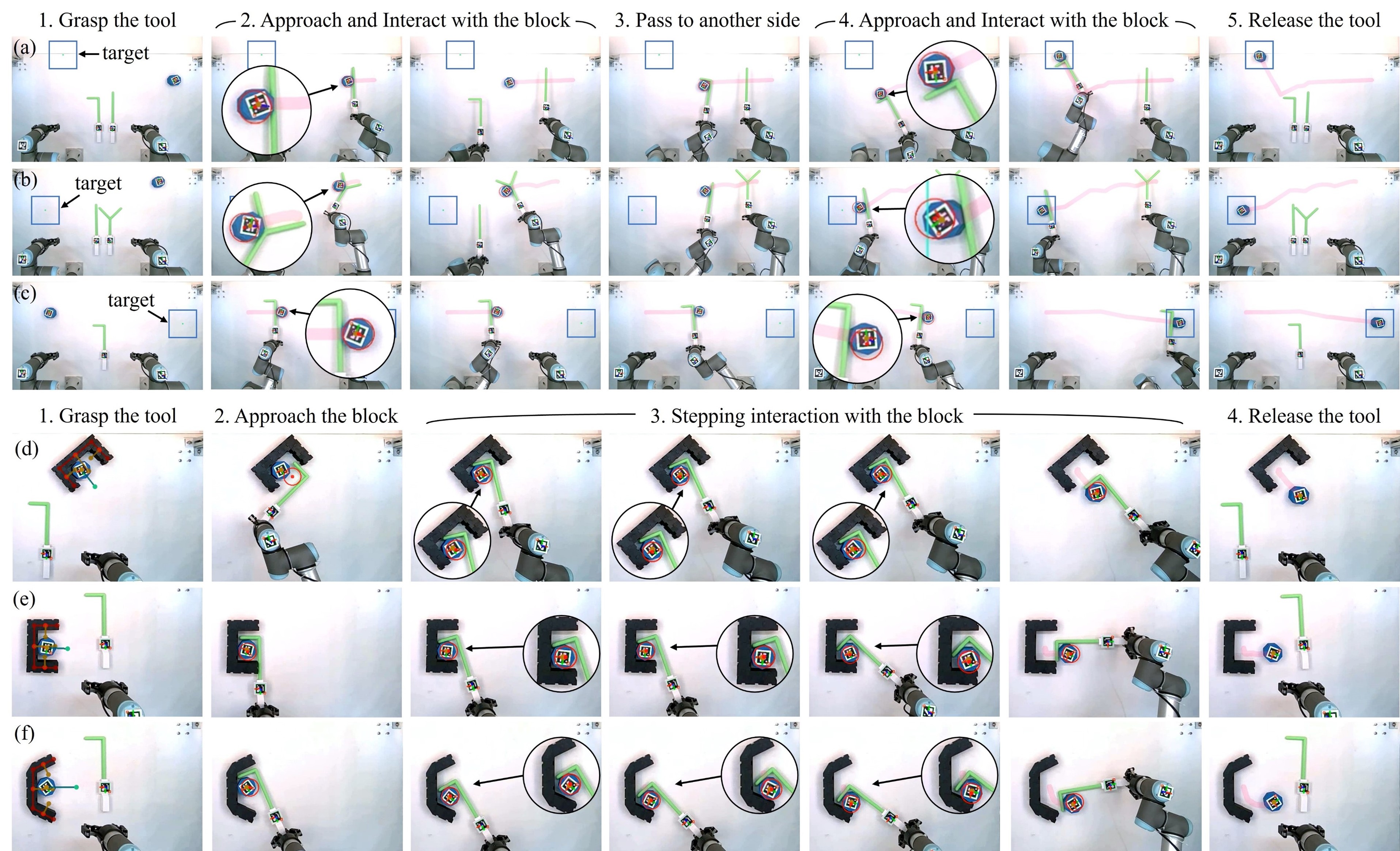}
    \caption{Long-horizon task: moving the manipulandum from (a) far right to far left with a hook and a stick, (b) far top right to far left with a stick and a Y-shaped tool, (c) far left to far right with a hook; and (d)--(f) exit from a confined area with a stepping controller. The manipulandum trajectory is reflected in pink and the target is labelled with a blue square.}
    \label{fig:exp2}
\end{figure*}

We then evaluate the long-horizon task performance where the manipulandum has to travel from far right to far left, far right/left to top right/left, and vice versa. 
The long-horizon task is evaluated with multiple tool combinations. 
The system observes and generates a collaborative motion plan. In the experiment shown in Fig. \ref{fig:exp2}(a), the right and left arms pick up the stick and the hook respectively. The right arm uses the stick to push the manipulandum to the left side, allowing the left arm to continue the task. The robot leverages the advantage of the hook to drag the manipulandum closer to its working area and push the manipulandum to the desired location. 
In Fig. \ref{fig:exp2}(b), the right and left arms grasped the Y-shaped tool and the stick respectively. The right arm uses the tool to pass the manipulandum to the left. The left arm uses the stick to push the manipulandum to the target location.

\begin{figure}[t]
    \centering
    \includegraphics[width=1\linewidth]{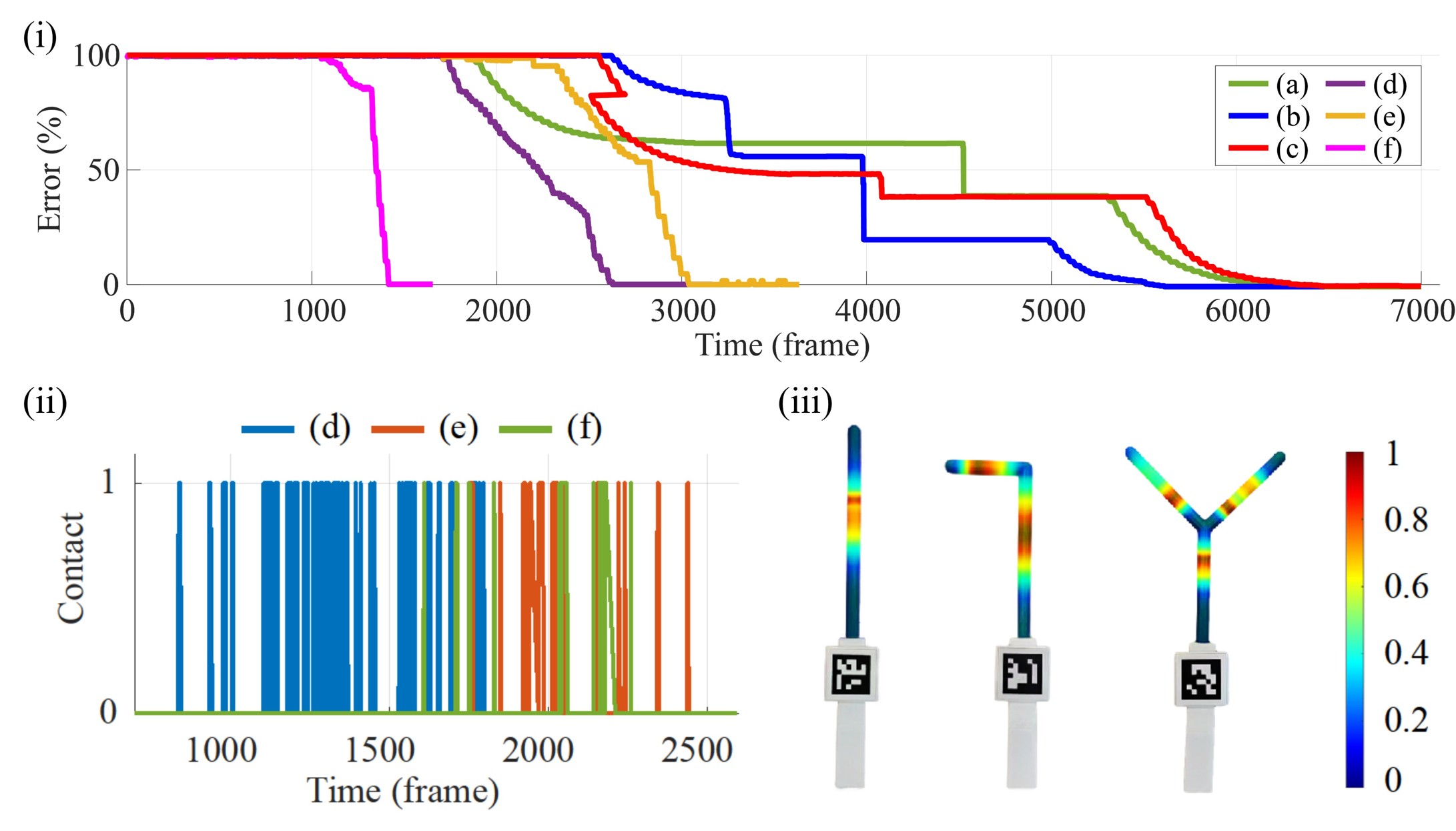}
    \caption{(i) Minimisation process of the error between the current object position and the target for the tasks shown in Fig. \ref{fig:exp2}. (ii) Stepping movement evolution of the change in contact between the manipulandum and the highest manoeuvrability point for the tasks shown in Fig. \ref{fig:exp2}(d)--(f). $1$ refers to in-contact and $0$ refers to no contact. (iii) Contact frequency of a segment side: regions depicted in deeper red indicate higher contact frequency with the manipulandum and a higher occurrence of affordance provision. (iv)--(v) Comparison of success rate and accuracy of tool manoeuvrability points under different state-of-the-art methodologies. FT states for fine-tuning, SRST states for a single-arm robot with a single tool, Dual refers to dual arms collaboration with two tools, and Sharing refers to tool-sharing collaboration. }
    \label{fig:exp2Error}
\end{figure}

\begin{table}[t]
\fontsize{10}{13}\selectfont
\caption{Manipulation Accuracy Across Subtasks for Different Tools}
\label{table:tbl_llm_success} 
\begin{center}
\begin{tabular}{p{2cm} m{1.7cm} m{1.7cm} m{1.7cm}}
\toprule
Subtask & Stick & Hook & Y-hook\\ \midrule
Grasp & 100\% & 100\% & 100\% \\ 
Approach & 96\% & 97\% & 95\% \\  
Interact & 91\% & 92\% & 92\% \\  
Pass & 92\% & 93\% & 93\% \\  
Release & 100\% & 100\% & 100\% \\ \bottomrule
\end{tabular}
\end{center}
\end{table}

\begin{figure}[t]
    \centering
    \includegraphics[width=1\linewidth]{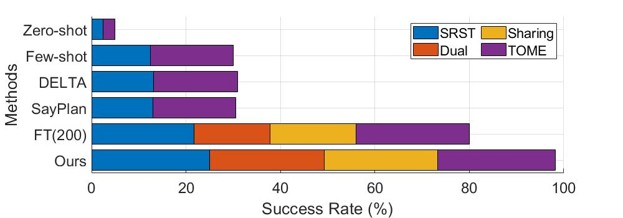}
    \caption{Comparison of success rate under different state-of-the-art methodologies. FT states for fine-tuning, SRST states for a single-arm robot with a single tool, Dual refers to dual arms collaboration with two tools, and Sharing refers to tool-sharing collaboration. }
    \label{fig:successRate_2}
\end{figure}

The long-horizon task performance is evaluated with the tool-sharing ability. Assuming there is only one tool available, it has to be shared among the dual-arm robot.  
Fig. \ref{fig:exp2}(c) demonstrates that the tool is passed to another arm once the manipulandum is pushed to the middle of the table.
The manipulandum is moved accurately to the target with motion-decomposed: `$\mathtt{grasp};$ 
$\mathtt{approach};$
$\mathtt{interact};$
$\mathtt{pass};$
$\mathtt{release};$
$\mathtt{grasp};$
$\mathtt{approach};$
$\mathtt{interact};$
$\mathtt{release}$'
where the left arm releases the tool once it is done and the right picks up the tool to continue moving the manipulandum. Though the hook is in a two-link geometry, the pushing is afforded by the right side of the tool (a single segment) with the highest manoeuvrability region.

The minimisation of the error between $\mathbf p^{\text{obj}}$ and $\mathbf p^{\text{target}}$ for each experiment is shown in Fig. \ref{fig:exp2Error}.
Similar to the single-arm robot with a single tool experiment, this long-horizon task also demonstrates the robustness of the proposed methodology such that the tasks are successfully decomposed into multiple collaborative subtasks, and the highest manoeuvrability region of the tool is leveraged in manipulandum manipulation.

To further evaluate subtask performance, Table \ref{table:tbl_llm_success} shows the manipulation accuracy, measured as the alignment between $\mathbf p^{\text{obj}}$ and the desired point for the \texttt{approach}, \texttt{interact}, and \texttt{pass} subtasks, and the success rate for the \texttt{grasp} and \texttt{release} subtasks across different tools. The results show 100\% success for \texttt{grasp} and \texttt{release}, while the rest of the subtasks maintain high accuracy above 90\% for all tools. These findings demonstrate that each subtask is executed reliably and that the proposed framework achieves robust performance across diverse tool geometries in long-horizon dual-arm tasks.

\subsection{Tool-Object Manipulation in Constrained Environments}
To further evaluate the performance of the model in application scenarios, different shapes of walls are constructed as shown in Fig. \ref{fig:exp2}(d)--(e).
Two walls are designed with 90-degree and 65-degree for the inner-angles. 
Maneuvering a hook within a confined space presents greater challenges compared to using a stick. Additionally, a Y-shaped hook proves unsuitable for dragging objects in tight quarters. Therefore, in this experimental study, we opt for a hook tool with a right arm to navigate effectively within the constrained environment.
Similar to the previous results, Fig. \ref{fig:exp2}(d)--(e) also implements the task planner successfully to decompose the task and applies the stepping controller for object manipulation. 
The tool first aligns its $\mathbf s^{\text{tip}}$ to the first segment of the wall and adopts the proposed non-prehensile stepping motion controller stated in Sec. \ref{envConstraints}. The manipulandum is dragged out from the confined area by alternating between the action of `repositioning' and `rotation-dragging'.

During the pulsing manipulation, the manipulandum maintains contact with the highest manoeuvrability region. 
The contact changes between the centre of the highest manoeuvrability region $\mathbf p_\ast$ with the manipulandum are visualized in Fig. \ref{fig:exp2Error}(ii). 
The error between the $\mathbf p^{\text{obj}}$ and the wall exit for each experiment is minimised with time, as shown in Fig. \ref{fig:exp2Error}.

\subsection{Comparison and Analysis}
We analyze the affordance utilization and provision for the selected tools by assessing the frequency of contact between the manipulandum and the tool segments.
In the majority of instances, the manipulandum interacts with the affordance primarily in the red region, as indicated in Fig. \ref{fig:exp2Error}(iii) and aligns closely with our proposed model.

We compare our system with other state-of-the-art methods. 
In terms of LLM-based task decomposition, we assess the success rates of our approach with zero-shot and few-shot learning methods \cite{brown2020language}, DELTA \cite{liu2024delta}, SayPlan \cite{rana2023sayplan}, and fine-tuning on a smaller dataset, as shown in Fig. \ref{fig:successRate_2}. In the comparison, zero-shot and few-shot learning refer to using prompts solely with a pre-trained model, rather than with a fine-tuned model. We consider task decomposition successful only if the output is optimal, with no unnecessary or redundant steps.

We observe that, under the same conditions, prompting (zero-shot and few-shot learning) is relatively unreliable, particularly in long-horizon tasks. This unreliability may stem from the insufficient number of manipulation examples provided in the prompt. Similarly, even when more information is given through domain knowledge and graphs \cite{liu2024delta, rana2023sayplan}, the LLM still struggles to generate a reasonable list for tasks involving both arms.

Fine-tuning a model with a smaller dataset (200 examples) yields acceptable results; however, it occasionally introduces unnecessary or infeasible steps in long-horizon tasks. In general, most methods demonstrate relatively positive outcomes in single-arm, single-tool tasks (SRST and TOME), likely due to the simplicity of these tasks. Specifically, the focus is on extracting the manipulandum from a constrained environment rather than aiming for a specific destination, and coordination between arms can be omitted. In summary, utilizing a larger dataset for fine-tuning results in enhanced task decomposition performance, leading to more consistent outcomes.

\begin{table}[t]
\small
\begin{center}
\caption{Success Rate Comparison in Task Decomposition}
\label{table:tbl_consistency} 
\scalebox{1.0}{  
\begin{tabular}{lcccc} \toprule
Methods & Our settings & NC1 & NC2 & Overall(\%)\\ \midrule
Zero-shot & 0.05 & - & - & 1.67\%\\ 
Few-shot & 0.30 & 0.24 & 0.13 & 22.3\%\\  
DELTA \cite{liu2024delta} & 0.31 & 0.26 & 0.14 & 23.7\%\\ 
SayPlan \cite{rana2023sayplan} & 0.31 & 0.25 & 0.14 & 23.0\%\\ 
PDDL \cite{jeon2022primitive} & 0.99 & 0.42 & 0.06 & 49.0\%\\ 
B. tree \cite{bagnell2012integrated} & 0.99 & 0.42 & 0.04 & 48.3\%\\ 
FT (200) & 0.83 & 0.77 & 0.69 & 76.3\%\\ 
Ours & 0.98 & 0.97 & 0.95 & 96.7\%\\ \bottomrule
\end{tabular}
}
\end{center}
\end{table}

To evaluate the generalizability of our framework against state-of-the-art methods, we conducted a comparative analysis of various approaches to robot task planning, focusing on their success rates in previously unseen task scenarios. The evaluated methods include Zero-shot learning, Few-shot learning, DELTA \cite{liu2024delta}, SayPlan \cite{rana2023sayplan}, planning domain definition language (PDDL) \cite{jeon2022primitive}, behaviour tree \cite{bagnell2012integrated}, and Fine-tuning with 200 data (FT 200), and our proposed approach. The results are summarized in Table \ref{table:tbl_consistency}.
Scenarios Evaluated: Our experiment settings: The position of the robot, tools, block, and target are based on our experiment settings; new case 1 (NC1): New language instruction with the positions are based on a slightly larger table and robot's workspace settings; new case 2 (NC2): New language instruction with the positions are based on a random-sized table and the robot's workspace settings.

\hl{
For a fair comparison of generalization capability, all baseline methods, including PDDL and Behavior Tree, were evaluated under a fixed configuration across all scenarios. Specifically, no manual retuning or reconfiguration (e.g., workspace bounds, distance thresholds, condition triggers, or collision margins) was performed for NC1 and NC2. Therefore, NC1 and NC2 are zero-shot transfer settings for these methods, where the task instructions and workspace scale change without updating the underlying symbolic rules or conditions.

In our experiment setting, most methods performed well, with PDDL and Behavior Trees achieving the highest initial success rates. However, their performance degraded significantly in NC1 and NC2, indicating limited generalizability under zero-shot transfer. Specifically, when the workspace was expanded or resized, the symbolic rules and absolute geometric thresholds used in PDDL and Behavior Trees, which were designed for the original workspace, were no longer applicable. For example, one condition stipulated that if the distance between the block and the right arm was below a fixed threshold, the right arm would initiate motion. Once the workspace scale changed, such conditions were no longer satisfied, resulting in no valid actions being triggered in certain situations.

A similar trend was observed in NC2, where our approach achieved a success rate of 0.95, while other methods showed substantially reduced performance. In contrast, fine-tuned learning-based methods maintain higher success rates because they can interpret task instructions and scene layouts directly from the prompt input and adjust their outputs accordingly. Overall, our proposed method demonstrates higher adaptability and generalization to unseen task instructions and workspace variations.
}

\begin{table}[t]
\centering
\small
\caption{Error in Tool Maneuverability Points}
\label{table:tbl_error_tool} 
\scalebox{1.0}{  
\begin{tabular}{l c c c} \toprule
Methods & Average(\%) & RMSE(\%) & MAE(\%)\\ \midrule
TVR & 38\% & 49\% & 23\% \\ 
Keypoint & 16\% & 18\% & 10\% \\  
Ours & 13\% & 15\% & 8\% \\ \bottomrule
\end{tabular}
}
\end{table}

\begin{figure}[t]
\centering
  \includegraphics[width=1\linewidth]{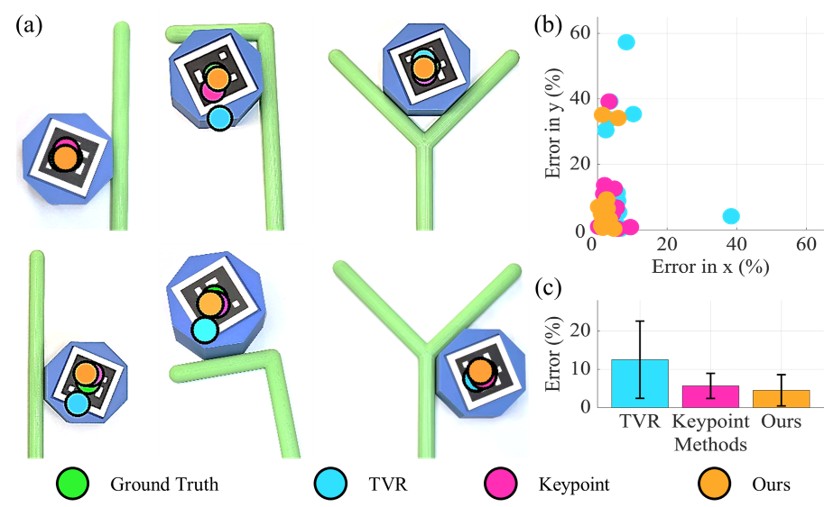}
  \caption{Comparison of tool maneuverability points under different state-of-the-art methodologies:
Green circles represent the ground truth, while blue, pink, and orange denote the computed results of the total variation regularization method, keypoint-inspired learning method, and our method respectively. (a) Differences visualization; 
(b) The average error between ground truth and computed results along the x and y axes in percentage;
(c) General differences in percentage.}
  \label{fig:comparison}

\end{figure}
\begin{figure}[t]
    \centering
    \includegraphics[width=1\linewidth]{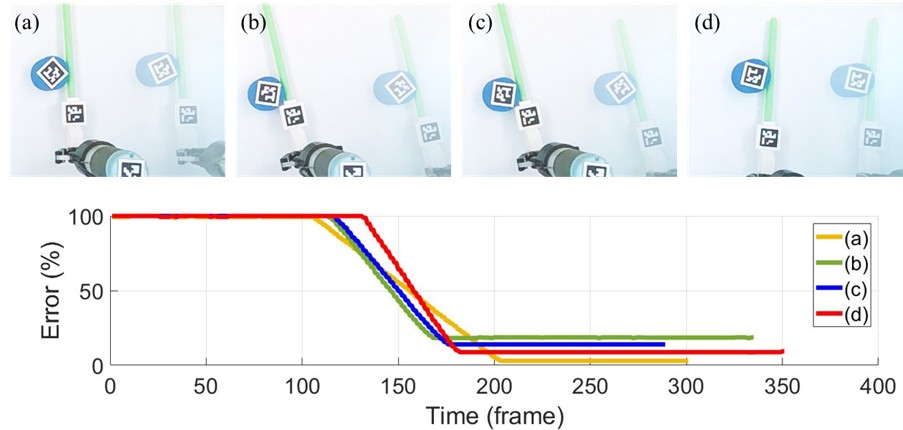}
    \caption{Component-wise ablation on point selection with a fixed subtask sequence. 
    Top row: (a) predefined fixed point with the proposed controller; 
    (b) endpoint-based selection; 
    (c) side-picker strategy; 
    (d) geometric-center selection. 
    Bottom row: evolution of the positional error between the object and the target for (a)--(d). 
    }
    \label{fig:compare_4_all}

\end{figure}

\begin{figure}[t]
    \centering
    \includegraphics[width=1\linewidth]{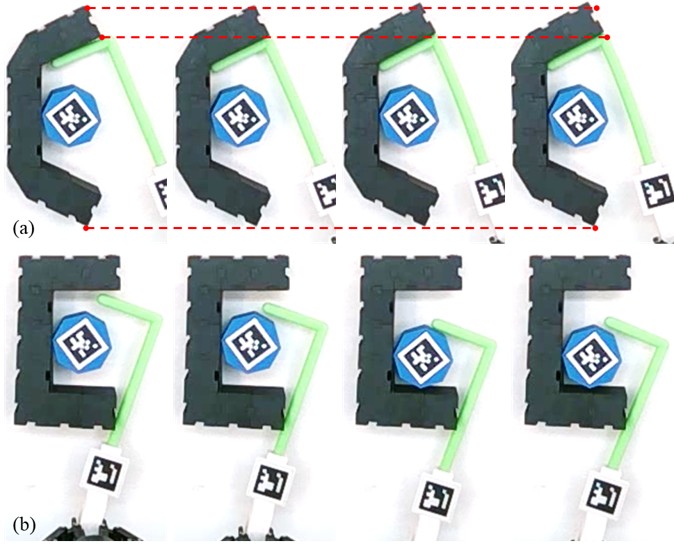}
    \caption{Results obtained using straight-line and basic position control in constrained environment settings. Structural deformation of the tools and walls is observed. The red dashed lines denote the reference geometry from the leftmost frame, illustrating the displacement of the wall relative to its initial configuration.}
    \label{fig:cat_add}

\end{figure}

We assess the tool analysis method by identifying the highest manoeuvrability point across 32 tool images, with the results outlined in Table \ref{table:tbl_error_tool} and Fig. \ref{fig:comparison}. 
The center of the manipulated manipulandum is taken as the ground truth. 
For our analysis, we consider the average error, root mean square error (RMSE), and mean absolute error (MAE) as the key metrics. 
The results are visualized in Fig. \ref{fig:comparison}, showcasing the differences between the ground truth and the computed results under various methodologies.
In the comparison, we observe that the total variation regularization (TVR) method \cite{pragliola2023and} had a relatively higher difference from the ground truth. The keypoint-inspired learning approach (similar to \cite{manuelli2019kpam}) yields comparable results to our method. However, the keypoint approach requires manual labelling of large amounts of data and model training, and its accuracy is highly dependent on the quality of the dataset.
As shown in Fig. \ref{fig:comparison}, both the keypoint and our methods had lower errors along the x-axis than the y-axis. Overall, both achieved relatively lower errors than the TVR method. Yet, in general, our proposed method demonstrated more stable performance and higher accuracy in terms of manoeuvrability computation.

\hl{
The proposed framework integrates LLM-based task decomposition, manoeuvrability-driven point selection, and a non-prehensile motion controller. To disentangle the contribution of individual components, we conduct a targeted ablation study by bypassing the LLM-based task decomposition and fixing the same subtask sequence across all trials. This allows us to isolate the effects of point selection and motion control on tool-object manipulation performance.

As shown in \mbox{Fig.\ref{fig:compare_4_all}}(a), we first replace the manoeuvrability-based point selection with a predefined fixed contact point while retaining the proposed motion controller. We then evaluate several baseline point-selection strategies: (b) selecting a point at a fixed distance from the tool endpoint, (c) a side-picker strategy that only enforces the object to remain on the correct side of the tool, and (d) selecting the geometric center of the tool.

The manipulation outcomes and the corresponding evolution of the position error between the object and the target are visualized in \mbox{Fig.\ref{fig:compare_4_all}}. With a fixed point, the controller can still manipulate the object stably and guide it toward the target. In contrast, the endpoint-based and side-picker strategies result in significant object slipping along the tool, leading to large final errors, as shown in \mbox{Fig.\ref{fig:compare_4_all}}(b,c). The geometric-center strategy yields behavior closer to the fixed-point baseline but still exhibits noticeable drift from the tool center.
These results indicate that effective point selection and a manoeuvrability-based controller are critical for stable non-prehensile manipulation.

We additionally evaluated simpler motion-control baselines in constrained environments, including straight-line pushing and basic position control. In these settings, such controllers frequently failed to maintain stable contact and resulted in collisions with surrounding boundaries, leading to task failure and potential damage to the environment and to the tool (see \mbox{Fig. \ref{fig:cat_add}}. These failures are primarily due to the strong reliance of simple controllers on accurately modeled workspace geometry and precise contact conditions. In contrast, the proposed stepping and rotation-dragging controller adapts its motion incrementally based on visual feedback, enabling safer and more robust interaction under spatial constraints without requiring an explicit or highly accurate environment model.
}

\section{Discussion and Conclusion}\label{sec:discussionConclusion}
In this paper, we present a new manoeuvrability-driven approach for tool-object manipulation. The LLM is integrated for task decomposition, generating collaborative motion sequences for a dual-arm robot system. A compact geometrical-based affordance model for describing the potential functionality and computing the highest manoeuvrability region of a tool is developed. A non-prehensile motion controller is developed for TOM, utilizing a stepping controller for incremental manipulation within a constrained environment. Experimental results are reported and analysed for the proposed methodology validation. 
We illustrate the performance of the proposed methods in the accompanying video \url{https://vimeo.com/917120431}. 

\subsection{Discussion}
Our method introduces a new affordance and manoeuvrability paradigm for tool-based object manipulation. To improve performance, the framework is separated into task decomposition and analytical motion models. This modular design allows the LLM to handle high-level planning using cloud computation, while the local computer executes physically grounded analytical models for low-level motion. 
In addition, the non-prehensile stepping controller enables incremental manipulation of objects in constrained environments. Unlike approaches that require computing an optimal trajectory for dragging the object out of a confined space, this method performs iterative small adjustments, alternating between tool repositioning and incremental rotation-dragging of the object until it is fully extracted. This strategy allows stable and precise manipulation in highly restricted areas, inspired by animal tool-use behaviors, and supports real-time execution without the need for high-end local GPU resources.

However, the method has limitations. The LLM may occasionally generate infeasible plans, which can lead to inappropriate motions.
To improve generalization and enhance transferability to unseen scenarios, future work will explore alternative strategies such as domain adaptation and transfer learning.
Moreover, the current affordance model presents promising results with simple geometrical shapes. Dynamic shapes like deformable objects may be complicated to perform accurate modelling. 
Manoeuvrability computation can also be affected by unstable illumination, low contrast, or large height differences between objects. In our experiments, these challenges were mitigated using ArUco markers for real-time tracking.

\subsection{Conclusion}
This work presents a manoeuvrability-driven framework for tool-object manipulation that integrates LLM-based task decomposition, a geometrical affordance model, and a non-prehensile stepping controller for incremental manipulation in constrained environments. Experimental validation demonstrates the effectiveness of this approach for collaborative dual-arm tasks and various tool configurations.

For future work, we would like to extend our method to deal with multiple object transportation and manipulation with tools. We would also like to perform deformable object manipulation, for example, the case of manipulating objects with ropes or fabrics. 
Also, we would like to test the performance of our controller but using other models. For that, the stability of the controller might be needed. We encourage readers to work on this open problem.

\bibliographystyle{elsarticle-num}
\bibliography{main.bib}

\end{document}